\documentclass{article} %
\usepackage[final]{colm2026_conference}

\usepackage{amsmath,amsfonts,bm}

\newcommand{\pmerr}[1]{\hspace{0.2em}{\scriptsize$\pm$\,#1}}

\def\eqref#1{equation~\ref{#1}}

\def\1{\bm{1}}

\DeclareMathAlphabet{\mathsfit}{\encodingdefault}{\sfdefault}{m}{sl}
\SetMathAlphabet{\mathsfit}{bold}{\encodingdefault}{\sfdefault}{bx}{n}

\usepackage{microtype}
\usepackage{hyperref}
\usepackage{url}
\usepackage{booktabs}
\usepackage{graphicx}
\usepackage{algorithm}
\usepackage{algorithmic}
\usepackage{multirow}
\usepackage{subcaption}
\usepackage{paralist}
\usepackage{bbm}
\usepackage[most]{tcolorbox}

\usepackage{xcolor}
\usepackage[normalem]{ulem}
\usepackage{fontawesome5}

\newtcolorbox{promptbox}[1]{
    colback=gray!5,
    colframe=gray!80!black,
    fonttitle=\bfseries\sffamily,
    title=#1,
    sharp corners,
    boxrule=0.5pt,
    breakable,
    enhanced,
    fontupper=\ttfamily\small %
}

\usepackage{lineno}

\newcommand{\sysname}{\textbf{\texttt{OpenStamp}} }
\newcommand{\sysnamenospace}{\textbf{\texttt{OpenStamp}}}
\newif\ifshowcomments
\showcommentstrue

\ifshowcomments
    \newcommand{\dd}[1]{\textcolor{blue}{\textbf{\small [#1 -- Danish]}}}
    \newcommand{\ddc}[2]{\textcolor{red}{\textbf{\small{\st{#1} #2}}}}

    \newcommand{\sr}[1]{\textcolor{magenta}{\textbf{\small [SR: #1]}}}
    \newcommand{\src}[2]{\textcolor{magenta}{\textbf{\small{\st{#1} #2}}}}

    \newcommand{\mb}[1]{\textcolor{green}{\textbf{\small [MB: #1]}}}
    \newcommand{\mbc}[2]{\textcolor{green}{\textbf{\small{\st{#1} #2}}}}
\else
    \newcommand{\dd}[1]{}
    \newcommand{\ddc}[2]{}

    \newcommand{\sr}[1]{}
    \newcommand{\src}[2]{}

    \newcommand{\mb}[1]{}
    \newcommand{\mbc}[2]{}
\fi

\definecolor{darkblue}{rgb}{0, 0, 0.5}
\hypersetup{colorlinks=true, citecolor=darkblue, linkcolor=darkblue, urlcolor=darkblue}

\title{\sysnamenospace: A Watermark for Open-Source Language Models}

\author{Miroojin Bakshi \\
Indian Institute of Science \\
\texttt{miroojinb@iisc.ac.in} \\
\And
Saksham Rastogi\thanks{Work done at the Indian Institute of Science} \\
Carnegie Mellon University \\
\texttt{iitdsaksham@gmail.com} \\
\And
Danish Pruthi \\
Indian Institute of Science \\
\texttt{danishp@iisc.ac.in}
}

\begin{document}

\ifcolmsubmission
\linenumbers
\fi

\maketitle

\begin{abstract}

With the growing prevalence of large language model (LLM) generated content, 
watermarking is considered a promising approach 
for attributing text to LLMs and 
distinguishing it from human-written content. 
A prominent class of techniques embeds subtle but detectable signals in generated text by 
modifying token sampling probabilities. 
However, such methods are unsuitable for open-source models, where users have white-box access
and can easily disable watermarking during inference.
In this work, we introduce \sysnamenospace,
a watermarking technique that
encodes the watermarking logic 
directly into the model weights
by modifying only the final projection, or unembedding, layer.
Through experiments across two models,
we show that \sysname achieves superior detection performance,
with minimal degradation in model capabilities compared to prior methods.
The implanted watermark is explicitly designed, and empirically confirmed,
to be more robust to paraphrasing attacks and harder to scrub off through post-hoc
fine-tuning than prior open-source watermarks.
To enable developers to watermark their models, 
we release our code alongside watermarked versions of $4$ 
popular open-source models.

\center \faGithub~\href{https://github.com/mb-14/openstamp}{\texttt{https://github.com/mb-14/openstamp}}

\end{abstract}

\section{Introduction}

Large language models (LLMs) are capable of generating human-like text, which has led to their widespread adoption in various applications~\citep{brown2020language, chowdhery2022palm}. As these models become more prevalent,
there are looming
concerns
about potential misuse,
such as generating large-scale misinformation ~\citep{oviedo2023risks},
influencing public opinion ~\citep{panditharatne2023ai},
or orchestrating social engineering attacks ~\citep{grbic2023social}.
To address these concerns,
it is critical to develop methods for
distinguishing LLM-generated content from human-written text.
Such methods can also be used to
attribute the content to its source model,
thereby promoting transparency and accountability
in LLM use, especially for high-stakes applications. 

A promising way to detect model-generated content is to watermark the text by embedding subtle, imperceptible signals into the model's outputs.
To implant such signals, several prominent watermarking techniques either modify the next-token probabilities \citep{kirchenbauer2023watermark, liu2024adaptive} or constrain the sampling process during generation \citep{aaronson2023reform, kuditipudi2024robust, dathathri2024scalable}.
While effective, these decoding-based techniques are incompatible with requirements of open sourcing, where users have full control over the generation process and can easily disable the watermarking logic.
This motivates the need for techniques
that embed watermarking signals
directly into the weights of the model,
making the watermark harder to erase.

Few recent works explore watermarking approaches for open-sourced LLMs.
A notable approach proposes to
distill a student model
using the outputs of the
watermarked teacher model~\citep{gu2023learnability}.
However, this method typically requires considerable amounts of training data and computational resources.
\citet{christ2024provably} instead add Gaussian noise to the final layer's bias vector to steer generation toward a fixed, context-independent set of tokens. This makes the scheme analogous to a fixed green list, which can be reverse-engineered and rendered ineffective \citep{jovanovic2024watermark,rastogi2024revisitingrobustness}.
Another concern is robustness to paraphrasing attacks \citep{krishna2023paraphrasing}, where an adversary rephrases the generated text to dilute, or remove, the watermarking signal. 
While some efforts have focused on 
improving robustness to paraphrasing \citep{liu2024a, ksemstamp2024}, 
none of these schemes work for open-source models.

\begin{figure*}[t]
    \centering
    \includegraphics[width=1.0\textwidth]{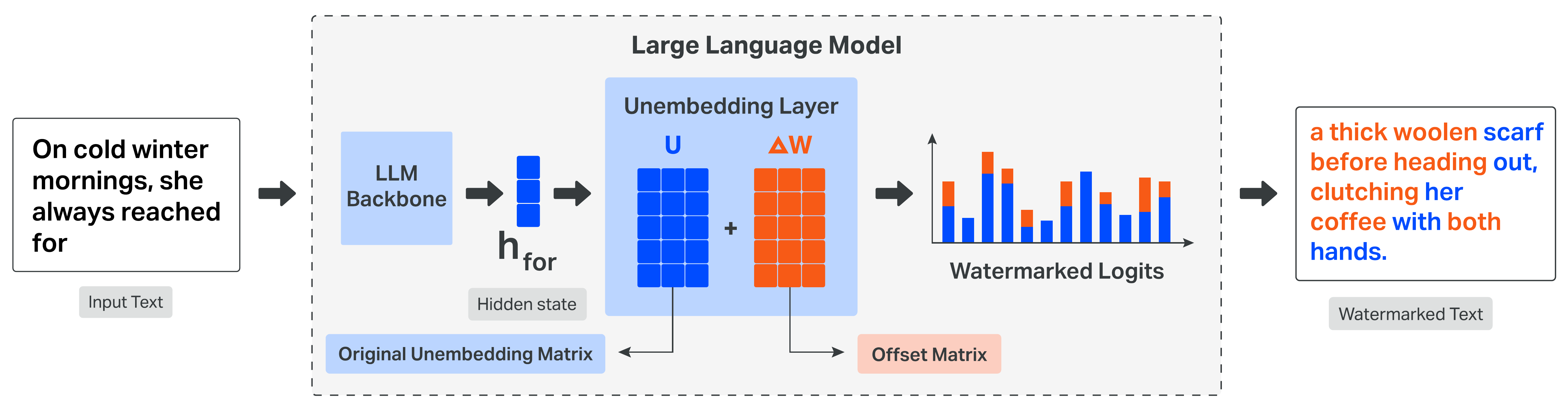}
    \caption{\textbf{Overview of \sysnamenospace, our watermarking method.}
        We add an offset matrix \( \Delta W\)
        to the unembedding layer's weights \(U\)
        to produce \textbf{watermark logits}
        that bias token sampling,
        favoring
        tokens with higher
        watermark logit values.
        The watermark can be
        detected using a \textbf{log-likelihood ratio (LLR)} based score
        which measures how much more
        likely the observed tokens are
        under the watermarked model
        compared to the unwatermarked model.
    }
    \label{fig:overview}
\end{figure*}

In this work, we introduce
\sysnamenospace,
an approach to watermarking open-source LLMs by modifying the \textit{unembedding layer}
weights of the model.
We add a carefully designed offset to these weights,
which bias the logits before generation and thereby
implant signals in the generated text. \sysname is conceptually
similar to prior logit-based watermarking approaches \citep{kirchenbauer2023watermark, liu2024a, liu2024adaptive}, which influence token selection by adding small, context-dependent biases to the logits at each decoding step.
However,
an important distinction is that
we embed the biasing logic directly into the unembedding layer's weights. 
Moreover, the offset is designed to produce consistent biases across semantically similar contexts, ensuring robustness against paraphrasing attacks. 
While conceptually similar to semantic watermarks \citep{semstamp2024, ksemstamp2024, liu2024a}, \sysname requires no auxiliary embedding model at generation or detection time, since this semantic robustness is baked into the model weights. 
Figure~\ref{fig:overview} shows a high-level overview of our watermarking method.
To detect watermarked text,
we compute a length normalized \textbf{log-likelihood ratio (LLR)} score,
which measures how much more
likely the
observed tokens are
under the watermarked model
compared to the unwatermarked model.

\sysname achieves near-perfect detection (TPR~$\geq$~99\% at 1\% FPR) with only minimal degradation in text quality,
outperforming other open-source watermarking methods
at comparable perplexity levels.
Our watermarking method approaches the Pareto frontier defined by
prominent decoding-based techniques. Furthermore, our watermarking signal is more difficult to erase via fine-tuning and offers improved robustness to paraphrasing attacks compared to existing approaches.
We believe that these strengths make our approach a
strong candidate for watermarking open-source models. To encourage model developers to watermark their models prior to release and to foster the development of new watermarking techniques for open-source models, we publicly release 
our code and watermarked versions of 
four popular 
open-source models 
created using \sysnamenospace. %

\section{Background and related work}
\label{sec:background}

\noindent \textbf{Watermark for large language models.}
Watermarking techniques for LLMs embed signals in model-generated text that remain imperceptible to human readers but can be algorithmically detected. A common approach alters the decoding process~\citep{kirchenbauer2023watermark,liu2024adaptive} by boosting a pseudorandom subset of tokens. Detection relies on statistical tests that exploit the resulting skew, for example by comparing the proportion of green tokens against the baseline expected under unwatermarked text. For a comprehensive survey, see~\cite{liang2024watermarkingtechniqueslargelanguage}.

\noindent \textbf{Paraphrasing robustness.}
Paraphrasing attacks pose a major challenge for watermark detection~\citep{krishna2023paraphrasing,kirchenbauer2024reliability}. By replacing or rearranging tokens, paraphrasing can dilute the statistical patterns that watermarking methods rely on for detection. Recent work has explored strategies to improve robustness, such as using semantic embeddings to guide the selection of green tokens~\citep{liu2024a, ksemstamp2024}, so that paraphrases tend to induce similar green lists. However, these methods depend on auxiliary models at inference time to produce semantic embeddings, making them unsuitable for open-source LLMs where users can easily bypass any watermarking logic that relies on external components.

\noindent \textbf{Watermarks for open-source LLMs.}
In open-source settings,
where users control the decoding process,
any decoding-time logic can be easily bypassed;
hence, the watermarking logic
must be embedded directly
into the model weights.
One approach uses distillation, training a student model on watermarked outputs from a decoding-based watermarked teacher, thereby enabling the student to generate watermarked text natively \citep{gu2023learnability}. While effective, this method requires substantial computational resources and training data. A reinforcement learning-based framework jointly optimizes an LLM and a paired detector to balance detectability and text quality \citep{xu2025learning}. A further line of work embeds a watermark into model weights by jointly fine-tuning LoRA adapters, optimizing for both coherence and detectability \citep{elhassan2025can}. Both methods are compute-intensive and rely on complex joint optimization objectives that are hard to tune.

Similar to our work are approaches that embed watermarking logic without requiring any LLM fine-tuning.
Unremovable~\citep{christ2024provably} introduces a fixed Gaussian perturbation to the final layer's bias vector,
steering generation towards positively biased tokens, similar to a fixed green list~\citep{unigram}.
Because the induced token preferences do not vary with context, this scheme is analogous to fixed green-list watermarks, which are known to be vulnerable to reverse-engineering attacks~\citep{jovanovic2024watermark,rastogi2024revisitingrobustness}.
GaussMark~\citep{block2025gaussmark} embeds a watermark by adding Gaussian noise to a subset of model weights, and detects it by analysing how the log-likelihood of a given text changes under the perturbed model.
However, GaussMark can be difficult to embed, as it requires careful selection
of the weight subsets and the noise strength
to balance text quality and detectability across models.

Despite recent innovations, a key limitation across existing work is the lack of robustness to post-hoc model modifications. This is a major challenge for watermarking open-source models, which are often updated through quantization, pruning, merging, or fine-tuning. \cite{gloaguen2025towards} find that no current methods remain detectable after such updates. We present complementary evidence on this limitation in Section~\ref{subsec:durability}.
Various recent approaches
underscore the inherent difficulty of watermarking in open-source LLMs.
Each method exhibits different limitations
across several axes: efficiency of integration,
detectability, robustness, and practicality.

\section{Methodology}
\label{sec:methodology}

\paragraph{Overview.}
This section presents our approach for watermarking language model outputs. We first introduce the model setup and notation (\S\ref{sec:preliminaries}), then describe how a modification to the unembedding layer can bias generation towards specific tokens to embed a watermark signal (\S\ref{subsec:watermarking-via-unembedding}). We outline key desiderata for effective watermarking and show how our design satisfies them (\S\ref{subsec:watermarking-logic}). Finally, we describe how to detect the watermark using a log-likelihood ratio based score (\S\ref{subsec:watermark-detection}).

\subsection{Preliminaries}
\label{sec:preliminaries}

Let a language model process a sequence of tokens drawn from a vocabulary \( \mathcal{V} \). Let \( x_t \in \mathcal{V} \) denote the token at generation step \( t \), and let \( x_{\leq t} = (x_1, \dots, x_t) \) denote the prefix up to step \( t \).
The model computes a hidden representation \( h_t = f(x_{\leq t}) \in \mathbb{R}^d \), where \( f: \mathcal{V}^* \rightarrow \mathbb{R}^d \) is the model's internal encoding function. The model then produces a logit vector
\[
v_t = U h_t \in \mathbb{R}^{|\mathcal{V}|},
\]
where \( U \in \mathbb{R}^{|\mathcal{V}| \times d} \) is the unembedding matrix. The model defines a categorical distribution \( p(x_{t+1} \mid x_{\leq t}) \) over the next token by applying a softmax over \( v_t \), where each component \( v_t^{(w)} \) corresponds to the logit value of token \( w \in \mathcal{V} \). Watermarking strategies typically modify \( v_t \) during generation to influence the next-token distribution.

\paragraph{KGW watermarking.}
In the KGW watermarking scheme ~\citep{kirchenbauer2023watermark}, the logit modification step is guided by a pseudorandom partition of the vocabulary. At each generation step \( t \), a pseudorandom function (PRF) selects a subset of tokens \( \mathcal{G}_t \subset \mathcal{V} \) called the \textit{green list}, based on the context 
\( x_{t-k:t} \) (where \( 0 < k \le t \))
and a secret key.
A parameter \( \gamma \) controls the 
fraction of tokens in the green list,
so that \( |\mathcal{G}_t| = \gamma |\mathcal{V}| \). 
The logits for tokens in the green list 
are boosted by a fixed hyperparameter \( \delta > 0 \):
\[
\hat{v}_t^{(w)} \;=\; v_t^{(w)} + \delta \cdot \mathbbm{1}\{ w \in \mathcal{G}_t \}.
\]

\subsection{Watermarking via unembedding matrix modification}
\label{subsec:watermarking-via-unembedding}

Modifying the unembedding matrix provides a direct mechanism to alter the logit vector produced at each generation step.
Concretely, we define a modified unembedding matrix
$\tilde{U} = U + \boldsymbol{\Delta W}$,
where we refer to $\Delta W \in \mathbb{R}^{|\mathcal{V}| \times d}$ as the \emph{offset matrix}. Applying this matrix to a hidden state $h_t$ yields the modified logit vector $\tilde{v}_t = \tilde{U} h_t = v_t + \boldsymbol{\Delta W h_t}$, where $\Delta W h_t$ is termed \emph{watermark logits}. These logits bias the output distribution during generation, favoring tokens with higher adjusted scores. This bias accumulates in the generated text, embedding a detectable watermark signal.

\paragraph{Desiderata.}
For the offset matrix to be effective, it should satisfy the following properties:
\begin{compactitem}
    \item \textbf{Detectability:} The watermark signal can be reliably identified in the generated text.
    \item \textbf{Controllability:} The balance between watermark strength and text quality is adjustable via hyperparameters.
    \item \textbf{Security:} The watermark is robust against reverse-engineering attacks.
    \item \textbf{Paraphrasing Robustness:} The watermark remains detectable even if the generated text is paraphrased.
\end{compactitem}

\subsection{Linearized green list biasing}
\label{subsec:watermarking-logic}

\paragraph{Conceptual overview.} 
We adapt the KGW scheme by modeling 
its green list selection as a linear transformation. 
While the PRF in KGW selects from a
combinatorially large space of lists
based on token prefixes, 
we simplify this mechanism by 
defining a finite set of $L$ candidate green lists. 
The offset matrix $\Delta W$ is then 
constructed to map hidden states to 
watermark logits that approximate
the logits associated with one of these $L$ lists.
Specifically, we decompose $\Delta W$ into three matrices:
\begin{equation}
\Delta W = G S P
\end{equation}
This decomposition realizes the watermarking logic through three sequential operations: 
(1) \textbf{semantic alignment} via $P$, 
which projects the raw hidden state $h_t$ 
reflects semantic similarity,
into a space where semantically similar sentences
have similar embeddings,
ensuring similar watermarking behavior 
for paraphrased sentences; 
(2) \textbf{list selection} via $S$, 
which maps the projected hidden 
state $Ph_t$ to a selector vector 
$s \in \mathbb{R}^L$ that acts 
as a soft selector over the $L$ 
candidate green lists; and (3) \textbf{logit biasing} via $G$, which maps the selector vector into watermark logits that encode the chosen green list. We describe the construction of each matrix in detail below.

\paragraph{Projection matrix \(P\).}
To ensure \textbf{paraphrasing robustness}, the watermarking logic must behave consistently across hidden states corresponding to semantically equivalent sentences. However, standard LLM hidden states are optimized for next-token prediction and may not cluster according to semantic similarity. To address this, we introduce $P \in \mathbb{R}^{d \times d}$ to map $h_t$ into a space where geometric proximity reflects semantic similarity. We train $P$ on a paired dataset of hidden states and high-quality sentence embeddings $\{(h_i, e_i)\}_{i=1}^N$ by minimizing the Kullback-Leibler (KL) divergence between their similarity distributions within a batch $\mathcal{B}$:
\begin{equation} \label{eq:projection-loss}
\mathcal{L}(P) = \frac{1}{|\mathcal{B}|} \sum_{i \in \mathcal{B}} D_{KL} ( \text{softmax} ( \mathbf{s}^P_i / T_{\mathrm{sim}} ) \| \, \text{softmax} ( \mathbf{s}^e_i / T_{\mathrm{sim}} ) ),
\end{equation}
where $T_{\mathrm{sim}} > 0$ is a temperature hyperparameter and the similarity vectors $\mathbf{s}^P_i, \mathbf{s}^e_i \in \mathbb{R}^{|\mathcal{B}|}$ represent the cosine similarities to all other elements in the batch:
\begin{equation*}
[\mathbf{s}^P_i]_j = \cos(Ph_i, Ph_j), \quad [\mathbf{s}^e_i]_j = \cos(e_i, e_j) \quad \forall j \in \mathcal{B}.
\end{equation*}
Minimizing this loss encourages the projected states to inherit the geometric structure of the target embedding space. Training details are provided in Appendix~\ref{appendix:projection-training}.

\paragraph{Selector matrix \(S\).}
The selector matrix \(S \in \mathbb{R}^{L \times d}\) maps the projected hidden state to a selection vector $s = S P h_t \in \mathbb{R}^L$. This vector acts as a \emph{soft selector} over $L$ candidate green lists. To train $S$, we partition the projected hidden states into $L$ clusters using $k$-means. This gives us a training dataset \(\mathcal{D}_{\text{train}} = \{(Ph_i, y_i)\}_{i=1}^N\), where \(y_i\) is the one-hot encoding of the cluster label for \(Ph_i\). We then learn $S$ by solving the following ridge regression problem:
\begin{equation*} \label{eq:regression}
    \min_S \sum_{(Ph_i, y_i) \in \mathcal{D}_{\text{train}}} \left\| SPh_i - y_i \right\|^2 + \lambda \| S \|_F^2,
\end{equation*}
where \(\lambda > 0\) is a regularization parameter. See Appendix~\ref{appendix:selector-training} for training details.

\paragraph{Green list matrix \(G\).}
The matrix \(G \in \mathbb{R}^{|\mathcal{V}| \times L}\) encodes \(L\) candidate green lists as its columns such that \(G s\) produces the desired watermark logits. For each column \(l \in \{1, \dots, L\}\), the corresponding green list \(\mathcal{G}_l \subset \mathcal{V}\) is defined by a PRF:
\[
\mathcal{G}_l = \{\, i \mid \mathrm{PRF}(\texttt{seed}, l, i) < \gamma \,\},
\]
where \(\texttt{seed}\) is the secret key shared between watermark generation and detection, and \(\mathrm{PRF}(\cdot)\) maps the token index \(i\) to a value in \([0,1]\). 
Each entry of \(G\) is then given by
\begin{equation*}
    G_{i,l} = \delta \cdot \mathbbm{1}\{ i \in \mathcal{G}_l \}.
\end{equation*}
The hyperparameters \(\gamma\) and \(\delta\) have the same interpretation as in the KGW watermarking scheme, offering \textbf{controllability} over the watermark's strength and impact on text quality.

\paragraph{Variability.}
By dynamically selecting among multiple green lists based on the context, our method maintains variability in the watermark logits, making it difficult to reverse-engineer. We provide supporting evidence of this variability in Appendix~\ref{appendix:variability-watermark-logits}.

\paragraph{Approximating one-hot selectors.}
Since our method operates on continuous hidden states, the selector matrix \(S\) yields soft, continuous selectors instead of exact one-hot vectors. As a result, multiple green lists can be partially activated at once leading to weaker alignment with the intended green-list behavior. Nonetheless, detection performance remains strong. We analyze this effect and its influence on detection in Appendix~\ref{appendix:understanding-watermark-behavior}.

\subsection{Detection via log-likelihood ratio}
\label{subsec:watermark-detection}

To detect the presence of a watermark, we compute a \textbf{length-normalized log-likelihood ratio (LLR)} for the given sequence using the watermarked 
and original unwatermarked model, specifically:
\begin{equation} \label{eq:llr}
    \text{LLR}(x) = \frac{1}{T} \sum_{t=1}^{T} \log \frac{p_{\text{wm}}(x_t \mid x_{<t})}{p_{\text{orig}}(x_t \mid x_{<t})},
\end{equation}
where the probabilities
are defined via softmax
over logits from the
original unwatermarked model, $p_{\text{orig}}$
and it's watermarked version, $p_{\text{wm}}$.

\paragraph{Intuition.}
The \textbf{LLR} score measures how much more likely a given token sequence is under the watermarked model than under the original model. Because the watermark logits systematically bias sampling towards certain favored tokens, generations from the watermarked model tend to accumulate higher LLR values, making the watermark \textbf{detectable} in practice. Length normalization ensures that a single detection threshold can be applied consistently across sequences of different lengths. 
While the conditional probabilities in Equation~\ref{eq:llr} formally condition on the full history, detection remains robust using only partial prefixes that exclude the generation prompt 
(See Appendix~\ref{appendix:prompt-independence} for details). \emph{Crucially, this means the detection process does not require knowledge of the specific prompt used to generate the text.} Accordingly, all results reported in the main paper are computed without assuming access to the prompt used to generate the watermarked text.

Unlike frequency-based detection methods~\citep{kirchenbauer2023watermark}, 
our LLR score does not constitute a formal statistical test.
Frequency-based detectors typically have a known expected green-token count on unwatermarked text and thus a calibrated $p$-value, regardless of whether the text is
human-written or produced by another model.
In contrast, our LLR compares the likelihoods assigned by two fixed models, so its typical value on
unwatermarked text shifts with the source. As a result, there is no single null distribution from which to compute a universally valid $p$-value~\citep{block2025gaussmark}.
Nevertheless, because the LLR uses the full token probability distribution and naturally captures
the \emph{mixture} of green lists produced by our watermarking method, it preserves signal that frequency-based tests lose by relying only on discrete counts (See Appendix~\ref{appendix:freq-ablation} for comparisons).

\begin{figure*}[t]
  \centering
  \includegraphics[width=0.79\textwidth]{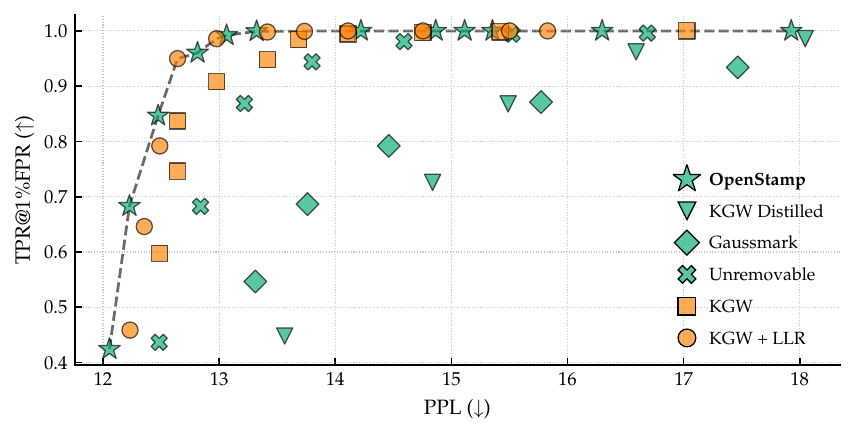}
  \caption{
    \textbf{Trade-off between detectability and text quality for various watermarking methods on \texttt{Llama-2-7B}.}
    \sysname is on the Pareto frontier and outperforms all open-source compatible methods (shown in green) by a substantial margin. Vanilla KGW performs slightly worse than KGW + LLR, highlighting the effectiveness of the LLR detection method.
  }
  \label{fig:ppl_tpr_plot}
\end{figure*}

\paragraph{Detection protocol.}
The developer publicly releases a 
watermarked model with the modified 
unembedding matrix \( \tilde{U} = U + \Delta W \), 
while keeping the original 
unembedding matrix private.
The developer can then expose detection as an API.
This prevents an adversary from 
recovering the watermark offset \(\Delta W = \tilde{U} - U\).
This also means that the method 
is not applicable to models whose 
original weights are already public. 
Any text generated using the 
public model can be detected 
using Algorithm~\ref{alg:wm_detection}.
This protocol assumes white-box access 
to the model weights in order 
to evaluate log-likelihoods.

\begin{table}[ht]
    \centering
    \small
    \setlength{\tabcolsep}{2.8pt}
    \begin{tabular}{llcccccc}
        \toprule
        \multirow{2}{*}{\textbf{Method}} & \multirow{2}{*}{\textbf{Model}} & \multicolumn{2}{c}{\textbf{ArXiv}} & \multicolumn{2}{c}{\textbf{BookSum}} & \multicolumn{2}{c}{\textbf{Wikipedia}} \\
        \cmidrule(lr){3-4} \cmidrule(lr){5-6} \cmidrule(lr){7-8}
        & & \textbf{TPR} & \textbf{PPL} & \textbf{TPR} & \textbf{PPL} & \textbf{TPR} & \textbf{PPL} \\
        \midrule
        \multirow{2}{*}{\textbf{KGW}$^\dagger$}
        & \texttt{Llama-2-7B} & 0.994\pmerr{0.005} & 32.3\pmerr{1.8} & 0.998\pmerr{0.002} & 24.1\pmerr{1.4} & 0.914\pmerr{0.036} & 13.8\pmerr{0.6} \\
        & \texttt{Mistral-7B} & 0.998\pmerr{0.002} & 30.2\pmerr{2.4} & 0.998\pmerr{0.002} & 24.6\pmerr{0.7} & 0.911\pmerr{0.092} & 13.2\pmerr{0.6} \\
        \cmidrule{2-8}
        \multirow{2}{*}{\textbf{KGW + LLR}$^\dagger$}
        & \texttt{Llama-2-7B} & 0.994\pmerr{0.010} & 32.0\pmerr{1.1} & 1.000\pmerr{0.001} & 23.2\pmerr{0.4} & 0.996\pmerr{0.004} & 12.7\pmerr{0.4} \\
        & \texttt{Mistral-7B} & 1.000\pmerr{0.000} & 29.2\pmerr{1.2} & 1.000\pmerr{0.001} & 23.6\pmerr{0.5} & 0.992\pmerr{0.009} & 13.0\pmerr{1.0} \\
        \midrule
        \multirow{2}{*}{\shortstack[l]{\textbf{KGW}\\\textbf{Distilled}}}
        & \texttt{Llama-2-7B} & 0.886\pmerr{0.076} & 43.48\pmerr{0.7} & 0.927\pmerr{0.035} & 28.6\pmerr{2.4} & 0.813\pmerr{0.141} & 16.4\pmerr{1.7} \\
        & \texttt{Mistral-7B} & 0.983\pmerr{0.011} & 44.5\pmerr{4.6} & 0.998\pmerr{0.001} & 34.6\pmerr{2.1} & 0.878\pmerr{0.096} & 20.1\pmerr{0.8} \\
        \cmidrule{2-8}
        \multirow{2}{*}{\textbf{GaussMark}}
        & \texttt{Llama-2-7B} & 0.644\pmerr{0.278} & 40.0\pmerr{0.6} & 0.847\pmerr{0.076} & 30.4\pmerr{2.3} & 0.750\pmerr{0.027} & 17.2\pmerr{0.6} \\
        & \texttt{Mistral-7B} & 0.395\pmerr{0.126} & 33.3\pmerr{2.0} & 0.847\pmerr{0.125} & 26.6\pmerr{0.5} & 0.387\pmerr{0.199} & 14.3\pmerr{0.4} \\
        \cmidrule{2-8}
        \multirow{2}{*}{\textbf{Unremovable}}
        & \texttt{Llama-2-7B} & 0.997\pmerr{0.002} & 28.9\pmerr{1.2} & 0.997\pmerr{0.003} & 25.2\pmerr{3.3} & 0.871\pmerr{0.042} & 14.5\pmerr{1.6} \\
        & \texttt{Mistral-7B} & 0.987\pmerr{0.013} & 26.3\pmerr{2.7} & \textbf{0.995\pmerr{0.005}} & 24.7\pmerr{1.7} & 0.883\pmerr{0.081} & 13.3\pmerr{0.7} \\
        \cmidrule{2-8}
        \multirow{2}{*}{\textbf{\sysname}}
        & \texttt{Llama-2-7B} & \textbf{1.000\pmerr{0.000}} & 30.7\pmerr{2.3} & \textbf{0.999\pmerr{0.001}} & 25.6\pmerr{3.1} & \textbf{0.998\pmerr{0.002}} & 14.4\pmerr{0.9} \\
        & \texttt{Mistral-7B} & \textbf{1.000\pmerr{0.000}} & 29.3\pmerr{1.3} & 0.995\pmerr{0.002} & 25.5\pmerr{0.3} & \textbf{0.990\pmerr{0.004}} & 12.9\pmerr{0.6} \\
        \bottomrule
    \end{tabular}
    \caption{
        \textbf{Comparing different watermarking approaches across multiple datasets.}
        We report TPR@0.1\%FPR (labeled TPR) to measure detectability and perplexity (PPL) to characterize text quality. Bold values indicate the best TPR@0.1\%FPR at the lowest PPL for each dataset and model. \textbf{KGW}$^\dagger$ and \textbf{KGW + LLR}$^\dagger$ are decoding-based watermark methods included for reference. For both models, \sysname achieves near-perfect detection with PPL competitive with Unremovable and lower than GaussMark and KGW Distilled, demonstrating a superior tradeoff between text quality and detectability across the three datasets.
    }
    \label{tab:detect-tpr-ppl}
\end{table}

\section{Experimental setup}
\label{sec:experimental-setup}

We evaluate our watermarking method across four key axes: \textbf{(i)}~detection performance (\S~\ref{subsec:detection-performance}), \textbf{(ii)}~robustness to paraphrasing attacks (\S~\ref{subsec:paraphrasing}), \textbf{(iii)}~resistance to post-hoc model modifications (\S~\ref{subsec:durability}) and \textbf{(iv)}~impact on downstream tasks (\S~\ref{subsec:downstream}).
We conduct experiments across all four axes on \texttt{Llama-2-7B}~\citep{touvron2023llama} and \texttt{Mistral-7B}~\citep{jiang2023mistral7b}. To evaluate the generalizability of our method, we also report detection performance 
on four additional 
models: \texttt{phi-4}~\citep{phi4}, \texttt{Olmo-3-7B}~\citep{olmo2025olmo3}, \texttt{Qwen2.5-7B}~\citep{qwen2.5}, and \texttt{SmolLM2-1.7B}~\citep{allal2025smollm2smolgoesbig}, in Appendix~\ref{appendix:detection-other-models}.

\subsection{Baselines}
We compare our method against $3$ open-source-compatible watermarking approaches: 
Unremovable~\citep{christ2024provably}, GaussMark~\citep{block2025gaussmark}, 
and a distilled version of KGW~\citep{gu2023learnability}.\footnote{We omit RL-based watermarking~\citep{xu2025learning} as the model either produces repetitive sequences or the detection rate is unacceptably low (See Appendix~\ref{appendix:rl-watermarking-challenges} for details).} 
We also include KGW as a representative of conventional decoding-based techniques. Further, 
we evaluate KGW combined with the LLR-based detector (\S~\ref{subsec:watermark-detection}).
Details about hyperparameters 
are in Appendix~\ref{appendix:experimental-setup}.

\subsection{Evaluation protocol}
\label{subsec:evaluation-protocol}
\paragraph{Watermarked sample generation.}
We generate $500$ watermarked completions of $200$ tokens each, using $50$-token prompts sampled from the \texttt{RealNewsLike} subset of C4~\citep{raffel2020exploring}.
The corresponding unwatermarked completions are taken directly from the dataset as the next $200$ tokens after each prompt.
Detection is performed only on the continuation
following the prompt.
Watermarked generations are sampled using nucleus sampling with temperature $1.0$.
To assess generalizability, we also evaluate on prompts drawn from \texttt{ArXiv}~\citep{cohan2018discourse}, \texttt{BookSum}~\citep{kryscinski2022booksum}, and Wikipedia~\citep{wikidump2024}.

\paragraph{Metrics.}
We evaluate two properties: \textit{watermark detectability} and \textit{text quality}.
\textbf{Detectability} is measured by the AUROC and the true positive rate at a fixed false positive rate of 1\% (TPR@1\%FPR) and 0.1\% (TPR@0.1\%FPR). 
For each target FPR, 
we choose the detection threshold accordingly.
In applications such as plagiarism detection, 
where false positives carry a high cost, 
maintaining a low FPR is essential. 
\textbf{Text quality} is 
measured by the mean perplexity (PPL) 
of the watermarked samples, 
computed using \texttt{Llama-2-13B} as 
the oracle model. 
All metrics are averaged 
over three random seeds.

\section{Results}
\label{sec:experiments}

\subsection{Detection performance}
\label{subsec:detection-performance}

Figure~\ref{fig:ppl_tpr_plot} illustrates the tradeoff between text quality and detectability for various watermarking methods on \texttt{Llama-2-7B}. Each method's tradeoff curve is generated by varying a method-specific parameter that controls strength of the signal. Open-source methods are shown in green, while decoding-based methods are shown in orange.

\sysname achieves near-perfect detection (TPR~$\geq$~99.9\%) at a perplexity of approximately 13.3, outperforming the open-source baselines at comparable perplexity: Unremovable reaches about $87\%$ TPR@1\%FPR, while GaussMark and KGW Distilled reach only $45$--$55\%$. 
Furthermore, \sysnamenospace's superior detection power 
is consistent across multiple datasets 
even at 0.1\% FPR, 
as shown in Table~\ref{tab:detect-tpr-ppl}. 
Additionally, 
Appendix~\ref{appendix:ood-detection} reports results under 
significant distribution shift, including evaluating 
watermarking approaches for Japanese, mathematical web text, and python code.

The gap between KGW and KGW+LLR in Figure~\ref{fig:ppl_tpr_plot} shows that the LLR detector improves detectability.
Appendix~\ref{appendix:llr-baselines} reports the same improvement for the baseline methods when we replace each method's native detector with the LLR detector, highlighting the LLR detector's 
stronger detection capabilities.

\subsection{Robustness to paraphrasing attacks}
\label{subsec:paraphrasing}
To test robustness against paraphrasing attacks, we prompt an instruction-tuned LLM to paraphrase watermarked samples and measure the detectability of the paraphrased outputs. 
Details on the prompt and model used for paraphrasing are provided in Appendix~\ref{appendix:paraphrasing-attack}.
As a reference method designed for paraphrasing robustness, we include SIR~\citep{liu2024a}, which selects green tokens from a semantic embedding of the preceding context, so that paraphrases tend to induce similar green lists. To isolate the effect of the projection matrix $P$ in \sysname on paraphrasing robustness, we evaluate an ablation of our method without the projection matrix (\sysname w/o sem-align), where the selector matrix $S$ is trained on raw hidden states.

Table~\ref{tab:paraphrasing-robustness} shows \sysname achieves the highest TPR@1\%FPR on both models, demonstrating superior robustness to paraphrasing attacks. Ablating the projection matrix degrades performance, confirming its contribution in enhancing robustness to paraphrasing.

\begin{table}[h]
    \centering
    \small
    \begin{tabular}{lcccc}
        \toprule
        & \multicolumn{2}{c}{\textbf{\texttt{Llama-2-7B}}} & \multicolumn{2}{c}{\textbf{\texttt{Mistral-7B}}} \\
        \cmidrule(lr){2-3} \cmidrule(lr){4-5}
        \textbf{Method} & \textbf{AUROC} & \textbf{TPR@1\%FPR} & \textbf{AUROC} & \textbf{TPR@1\%FPR} \\
        \midrule
        SIR              & \text{0.973 $\pm$~0.007} & \text{0.753 $\pm$~0.035} &  \text{0.971 $\pm$~0.007}   &  \text{0.763 $\pm$~0.043}    \\
        \midrule
        KGW Distilled    & \text{0.950 $\pm$~0.004} & \text{0.602 $\pm$~0.021} &  \text{0.954 $\pm$~0.013}   &  \text{0.596 $\pm$~0.051} \\
        GaussMark        & \text{0.946 $\pm$~0.002} & \text{0.550 $\pm$~0.035} & \text{0.925 $\pm$~0.016} & \text{0.409 $\pm$~0.061}  \\
        Unremovable      & \text{0.966 $\pm$~0.009} & \text{0.724 $\pm$~0.098} & \text{0.966 $\pm$~0.008} & \text{0.715 $\pm$~0.010} \\
        \sysname w/o sem-align & \text{0.971 $\pm$~0.005} & \text{0.762 $\pm$~0.024} & \text{0.960 $\pm$~0.012}   & \text{0.732 $\pm$~0.038}   \\
        \sysnamenospace  & \textbf{0.990 $\pm$~0.003} & \textbf{0.907 $\pm$~0.003} & \textbf{0.978 $\pm$~0.005} & \textbf{0.791 $\pm$~0.013} \\
        \bottomrule
    \end{tabular}
    \caption{\textbf{Detectability of paraphrased text.} On both models, \sysname achieves the highest TPR@1\%FPR on paraphrased text, outperforming SIR and the open-source baselines. \sysname w/o sem-align degrades performance, confirming that the projection matrix contributes to robustness against paraphrasing attacks.}
    \label{tab:paraphrasing-robustness}
\end{table}

\subsection{Resistance to post-hoc model modifications}
\label{subsec:durability}

We assess how well open-source 
compatible watermarking methods 
can resist post-hoc model modifications,
ranging from adversarial fine-tuning explicitly 
aimed at erasing the watermark to 
naive modifications such as 
instruction fine-tuning and quantization 
that users commonly apply after release.
We first simulate an adversary attempting 
to erase the watermark from model weights 
by further fine-tuning 
on OpenWebText~\citep{Gokaslan2019OpenWeb} 
using LoRA~\citep{hu2022lora}. 
Full experimental details are provided in Appendix~\ref{appendix:finetune-details}. 
We measure the TPR@1\%FPR after 
every 500 fine-tuning steps up to 2,500 steps (2,500 steps comprise $\sim 60$ million tokens).
We discuss an additional 
attack setup in Appendix~\ref{appendix:additional-finetuning-attacks}.

\begin{figure}[t]
  \centering
  \begin{minipage}[t]{0.48\linewidth}
    \centering
    \includegraphics[width=\linewidth]{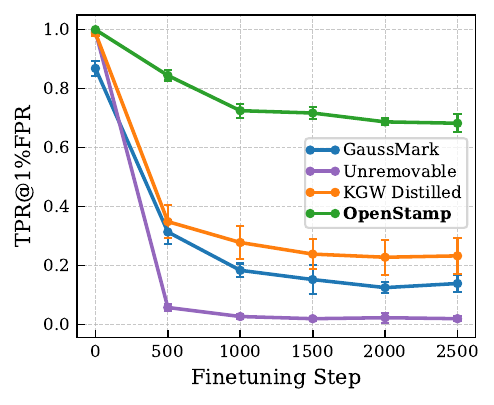}
    \caption{
      \textbf{\texttt{Llama-2-7B} detectability under post-hoc fine-tuning.}
      All methods degrade over time, but \sysname\ maintains higher detectability compared to GaussMark, KGW Distilled, and Unremovable. Note that 2,500 steps comprise $\sim 60$ million tokens.
    }
    \label{fig:durability}
  \end{minipage}\hfill
  \begin{minipage}[t]{0.48\linewidth}
    \centering
    \includegraphics[width=\linewidth]{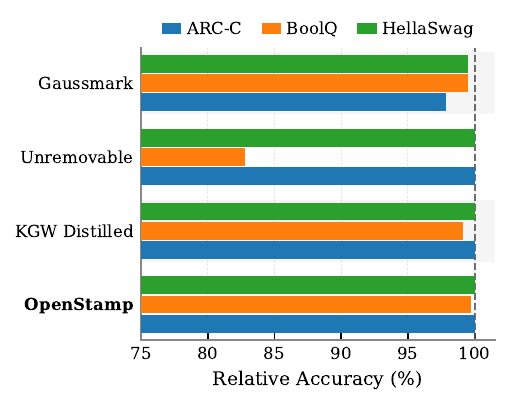}
    \caption{
      \textbf{Relative downstream task accuracy of watermarked \texttt{Llama-2-7B}.}
      Accuracy is shown as a percentage of the unwatermarked baseline (dashed line at 100\%). \sysname exhibits minimal degradation in accuracy across all benchmarks.
    }
    \label{fig:relative-accuracy-arc}
  \end{minipage}
\end{figure}

Figure~\ref{fig:durability} shows that while all watermarks degrade during \texttt{Llama-2-7B} fine-tuning, \sysname maintains superior detectability over GaussMark, KGW Distilled, and Unremovable. Similar trends hold for \texttt{Mistral-7B} (See Figure~\ref{fig:durability-mistral}).

\paragraph{Instruction fine-tuning.}
The OpenWebText experiment above simulates an adversary who continues fine-tuning on web text to scrub the watermark. Released checkpoints are commonly instruction-tuned. 
We evaluate this setting by fine-tuning watermarked models on the Alpaca instruction dataset~\citep{taori2023alpaca}. Training details are in Appendix~\ref{appendix:finetune-details}. To measure durability on both general-domain and instruction-domain outputs, we then evaluate detectability on RealNewsLike completions under the protocol in Section~\ref{subsec:evaluation-protocol}, and on instruction-following completions from AlpacaEval~\citep{li2023alpacaeval}.
As shown in Table~\ref{tab:instruction-finetuning}, detectability on RealNewsLike remains strong for all methods, with \sysname retaining the highest TPR@1\%FPR. On AlpacaEval, TPR@1\%FPR drops for every method, with Unremovable strongest and \sysname remaining competitive. Corresponding \texttt{Mistral-7B} results follow the same pattern (Appendix~\ref{appendix:mistral-additional-results}). These results are consistent with the findings of~\citet{gloaguen2025towards}: after instruction fine-tuning, the watermark can remain strong on general text while weakening on instruction-following outputs.
\begin{table}[h]
    \centering
    \small
    \setlength{\tabcolsep}{4pt}
    \begin{tabular}{lcccc}
        \toprule
        & \multicolumn{2}{c}{\textbf{RealNewsLike}} & \multicolumn{2}{c}{\textbf{AlpacaEval}} \\
        \cmidrule(lr){2-3} \cmidrule(lr){4-5}
        \textbf{Method} & \textbf{AUROC} & \textbf{TPR@1\%FPR} & \textbf{AUROC} & \textbf{TPR@1\%FPR} \\
        \midrule
        KGW Distilled & \text{0.996 $\pm$~0.002} & \text{0.952 $\pm$~0.006} & \text{0.634 $\pm$~0.206} & \text{0.090 $\pm$~0.135} \\
        GaussMark     & \text{0.980 $\pm$~0.003} & \text{0.775 $\pm$~0.014} & \text{0.885 $\pm$~0.027} & \text{0.301 $\pm$~0.092} \\
        Unremovable   & \text{0.996 $\pm$~0.001} & \text{0.965 $\pm$~0.010} & \textbf{0.984 $\pm$~0.005} & \textbf{0.707 $\pm$~0.101} \\
        \sysnamenospace & \textbf{1.000 $\pm$~0.000} & \textbf{1.000 $\pm$~0.000} & \text{0.977 $\pm$~0.005} & \text{0.699 $\pm$~0.038} \\
        \bottomrule
    \end{tabular}
    \caption{\textbf{\texttt{Llama-2-7B} detectability under instruction fine-tuning.} TPR@1\%FPR remains high on RealNewsLike but drops on AlpacaEval. Bold values mark the best in each column.}
    \label{tab:instruction-finetuning}
\end{table}

\paragraph{Quantization.}
Open-source models are commonly quantized after release to reduce memory and inference cost.
To assess watermark durability in this setting, we apply 4-bit NormalFloat (NF4)~\citep{dettmers2023qlora} or 8-bit integer (INT8)~\citep{dettmers2022llmint8} to watermarked models, generate RealNewsLike completions from these quantized models, and measure detectability.
Table~\ref{tab:quantization-robustness} shows that detectability remains high for \sysname and the baselines under both schemes, consistent with prior work finding that quantization has little effect on watermark detectability~\citep{gloaguen2025towards}.
\begin{table}[h]
    \centering
    \small
    \setlength{\tabcolsep}{4pt}
    \begin{tabular}{llccc}
        \toprule
        \textbf{Method} & \textbf{Model} & \textbf{Unquantized} & \textbf{NF4} & \textbf{INT8} \\
        \midrule
        \multirow{2}{*}{\sysnamenospace} & \texttt{Llama-2-7B} & 0.998 & 1.000 & 1.000 \\
                                            & \texttt{Mistral-7B} & 1.000 & 1.000 & 1.000 \\
        \midrule
        \multirow{2}{*}{KGW Distilled}      & \texttt{Llama-2-7B} & 0.986 & 0.976 & 0.988 \\
                                            & \texttt{Mistral-7B} & 0.990 & 1.000 & 0.994 \\
        \midrule
        \multirow{2}{*}{GaussMark}          & \texttt{Llama-2-7B} & 0.846 & 0.854 & 0.842 \\
                                            & \texttt{Mistral-7B} & 0.746 & 0.736 & 0.704 \\
        \midrule
        \multirow{2}{*}{Unremovable}        & \texttt{Llama-2-7B} & 0.994 & 0.994 & 0.993 \\
                                            & \texttt{Mistral-7B} & 0.996 & 0.990 & 0.995 \\
        \bottomrule
    \end{tabular}
    \caption{\textbf{Detectability under quantization.} TPR@1\%FPR remains high for \sysname and the baselines under both NF4 and INT8.}
    \label{tab:quantization-robustness}
\end{table}

\paragraph{Reinitializing and retraining the unembedding layer.}
An adversary may attempt to remove the watermark by reinitializing the unembedding layer and relearning the mapping from hidden states to token logits to restore fluent generation. Restoring this capability is non-trivial, since model developers train LLMs on large curated datasets and rely on complex pretraining and post-training pipelines.
To illustrate this difficulty, we reinitialized the unembedding layer and retrained only that layer on FineWeb~\citep{penedo2024fineweb}.
Table~\ref{tab:retrain-unembedding} reports perplexity after 2,500 and 25,000 training steps. Even after training on 800M tokens, perplexity remains far worse than the watermarked baseline, showing that recovering quality is non-trivial and making this attack computationally costly.
\begin{table}[h]
    \centering
    \small
    \begin{tabular}{lc}
        \toprule
        \textbf{Condition} & \textbf{PPL} \\
        \midrule
        Watermarked baseline & 15.1 \\
        After 2{,}500 steps (80M tokens) & 78{,}529.4 \\
        After 25{,}000 steps (800M tokens) & 351.7 \\
        \bottomrule
    \end{tabular}
    \caption{\textbf{Perplexity after retraining a reinitialized unembedding layer.} Only the unembedding layer is reinitialized and retrained on FineWeb; all other weights remain frozen.}
    \label{tab:retrain-unembedding}
\end{table}

\subsection{Impact on downstream task accuracy}
\label{subsec:downstream}

Since watermarking methods
alter the output distributions of LLMs,
it is essential to ensure
that the underlying model's capabilities are
not compromised.
While the results in Figure~\ref{fig:ppl_tpr_plot}
demonstrate minimal degradation in perplexity,
\citet{ajith-etal-2024-downstream} find that
perplexity measurements cannot
reliably predict the performance trade-offs
due to watermarking.
We therefore evaluate watermarked models using the Language Model Evaluation Harness~\citep{eval-harness}, measuring potential degradation across three benchmarks:
ARC-C~\citep{ARC-C}, BoolQ~\citep{boolq}, and
HellaSwag~\citep{zellers2019hellaswag}.

As shown in Figure~\ref{fig:relative-accuracy-arc}, \sysname causes minimal accuracy degradation on \texttt{Llama-2-7b} across all benchmarks. Notably, the results are statistically indistinguishable from the unwatermarked baseline, indicating no meaningful impact on 
downstream performance. Similar results are observed for \texttt{Mistral-7b} (see Figure~\ref{fig:relative-accuracy-mistral}).

\section{Limitations}
\label{subsec:limitations}

Our work has several important limitations.
First, detection requires a forward pass
through the model,
making it computationally expensive
compared to methods
that perform statistical tests on generated text
alone.
Second, our method assumes access to token probabilities
from the base model.
However, this is a realistic assumption as
model owners typically have white-box access to their models.
While our detection algorithm achieves stronger performance, it is not grounded in a statistical test and thus lacks a probabilistic interpretation, such as confidence levels or false positive rates, that help quantify detection uncertainty.
\section{Conclusion}
\label{sec:conclusion}

We proposed an effective approach
for watermarking open-source LLMs by embedding the watermarking logic directly into the unembedding layer weights. This design natively integrates the watermark into the generation process, eliminating the need for decoding-time interventions.
Our approach outperforms existing baselines in detection performance while preserving text quality. Furthermore, it demonstrates superior robustness against both paraphrasing and post-hoc fine-tuning, while remaining controllable and easy to integrate. Future work could explore advanced offset matrix designs to further improve this robustness, and enable detection methods with statistical guarantees.

\section*{Acknowledgements}

We thank the reviewers for their feedback. 
We are grateful to Krishna Pillutla for discussions that positively shaped our work.
Additionally, DP is grateful to the National Payments Corporation of India (NPCI), 
Schmidt Sciences through their AI2050 program  (Grant G-24-66186), and Google for generously supporting his group’s research.

\bibliography{colm2026_conference}
\bibliographystyle{colm2026_conference}

\appendix
\newpage
\section*{Appendices}

\section{Watermark detection via LLR score}
The LLR-based detection algorithm is formalized in Algorithm~\ref{alg:wm_detection}. Given a sequence \(x\), we first extract the hidden states from the LLM backbone. We then compute the log-likelihood of the sequence under both the original model and the watermarked model (which incorporates the offset matrix \(\Delta W\)). Finally, we calculate the length-normalized LLR score and compare it against a threshold \(\tau\) to determine if the sequence is watermarked.

\begin{algorithm}
    \caption{Watermark Detection via LLR Score}
    \label{alg:wm_detection}
    \begin{algorithmic}[1]
        \REQUIRE Sequence $x = (x_1, \dots, x_T)$, LLM backbone $f(\cdot)$, unembedding matrix $U$, offset matrix $\Delta W$, threshold $\tau$
        \ENSURE \texttt{True} if $x$ is watermarked; else \texttt{False}

        \textcolor{gray}{\textit{// Extract hidden states from the LLM}}
        \STATE $(h_1, \dots, h_T) \gets f(x)$

        \textcolor{gray}{\textit{// Compute log-likelihood of sequence under original model}}
        \STATE $\ell_{\text{orig}} \gets \sum_{t=1}^{T-1} \log \mathrm{softmax}(U h_t)[x_{t+1}]$

        \textcolor{gray}{\textit{// Compute log-likelihood under watermarked model}}
        \STATE $\ell_{\text{wm}} \gets \sum_{t=1}^{T-1} \log \mathrm{softmax}((U + \Delta W) h_t)[x_{t+1}]$

        \textcolor{gray}{\textit{// Compute length-normalized LLR score}}
        \STATE $\mathrm{LLR}(x) \gets (\ell_{\text{wm}} - \ell_{\text{orig}}) / (T - 1)$

        \RETURN $(\operatorname{LLR}(x) > \tau)$
    \end{algorithmic}
\end{algorithm}

\section{Mistral additional results}
\label{appendix:mistral-additional-results}
\begin{figure}[h]
  \centering
  \begin{minipage}[t]{0.48\linewidth}
    \centering
    \includegraphics[width=\linewidth]{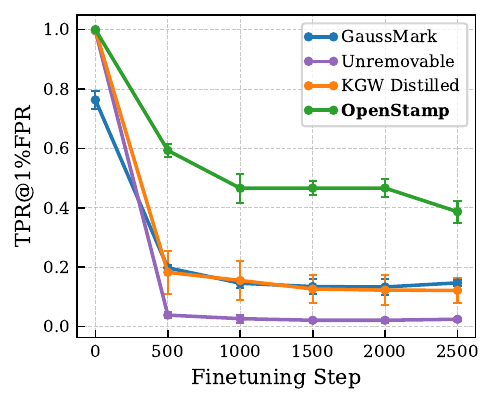}
    \caption{
      \textbf{\texttt{Mistral-7B} detectability under post-hoc fine-tuning.}
      All methods degrade over time, but \sysname\ maintains higher detectability compared to GaussMark, KGW Distilled, and Unremovable.
    }
    \label{fig:durability-mistral}
  \end{minipage}\hfill
  \begin{minipage}[t]{0.48\linewidth}
    \centering
    \includegraphics[width=\linewidth]{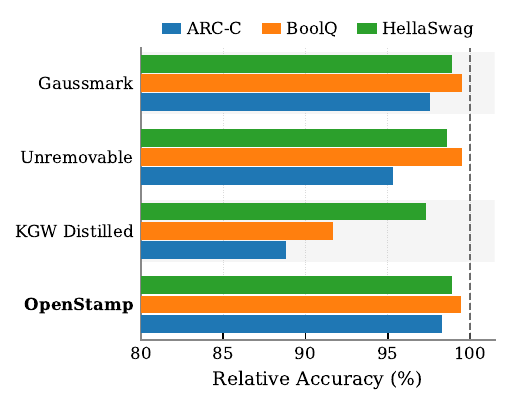}
    \caption{
      \textbf{Relative downstream task accuracy of watermarked \texttt{Mistral-7B}.}
      \sysname exhibits minimal degradation in accuracy across all benchmarks.
    }
    \label{fig:relative-accuracy-mistral}
  \end{minipage}
\end{figure}

We present additional results on the \texttt{Mistral-7B} model across three axes: \textbf{robustness to post-hoc fine-tuning} (Figure~\ref{fig:durability-mistral}), \textbf{relative downstream task accuracy} (Figure~\ref{fig:relative-accuracy-mistral}), and \textbf{detectability after instruction fine-tuning} (Table~\ref{tab:instruction-finetuning-mistral}).
As in the \texttt{Llama-2-7B} setting (Table~\ref{tab:instruction-finetuning}), detectability on RealNewsLike remains high after Alpaca instruction fine-tuning, with \sysname retaining the strongest TPR@1\%FPR, while AlpacaEval scores are lower across methods.
\begin{table}[h]
    \centering
    \small
    \setlength{\tabcolsep}{4pt}
    \begin{tabular}{lcccc}
        \toprule
        & \multicolumn{2}{c}{\textbf{RealNewsLike}} & \multicolumn{2}{c}{\textbf{AlpacaEval}} \\
        \cmidrule(lr){2-3} \cmidrule(lr){4-5}
        \textbf{Method} & \textbf{AUROC} & \textbf{TPR@1\%FPR} & \textbf{AUROC} & \textbf{TPR@1\%FPR} \\
        \midrule
        KGW Distilled & \text{0.997 $\pm$~0.002} & \text{0.936 $\pm$~0.049} & \text{0.597 $\pm$~0.178} & \text{0.066 $\pm$~0.087} \\
        GaussMark     & \text{0.963 $\pm$~0.007} & \text{0.595 $\pm$~0.064} & \text{0.853 $\pm$~0.044} & \text{0.223 $\pm$~0.128} \\
        Unremovable   & \text{0.996 $\pm$~0.001} & \text{0.939 $\pm$~0.015} & \textbf{0.979 $\pm$~0.003} & \textbf{0.589 $\pm$~0.133} \\
        \sysnamenospace & \textbf{1.000 $\pm$~0.000} & \textbf{0.998 $\pm$~0.002} & \text{0.948 $\pm$~0.020} & \text{0.454 $\pm$~0.009} \\
        \bottomrule
    \end{tabular}
    \caption{\textbf{\texttt{Mistral-7B} detectability under instruction fine-tuning.} TPR@1\%FPR remains high on RealNewsLike but drops on AlpacaEval, matching Table~\ref{tab:instruction-finetuning}. Bold values mark the best in each column.}
    \label{tab:instruction-finetuning-mistral}
\end{table}

\section{Generalizability across LLMs}
\label{appendix:detection-other-models}
To assess the generalizability of our method across diverse LLMs, we evaluate its detection performance on four additional models: \texttt{phi-4}~\citep{phi4}, \texttt{Olmo-3-7B}~\citep{olmo2025olmo3}, \texttt{Qwen2.5-7B}~\citep{qwen2.5}, and \texttt{SmolLM2-1.7B}~\citep{allal2025smollm2smolgoesbig}. This selection covers a variety of model families, parameter scales, and training stages (pretrained vs.\ instruction-tuned). For all models, we apply same hyperparameters (\(\gamma = 0.25\), \(\delta = 1.0\)) and evaluate the detection performance using the same protocol as described in Section~\ref{subsec:evaluation-protocol}. Table~\ref{tab:open-source-model-robustness} shows that \sysname achieves perfect TPR@1\%FPR across all four models, demonstrating its robustness and generalizability. The watermarked perplexity (WM PPL) values are reasonably close to the unwatermarked perplexity (Baseline PPL), indicating that the watermark does not significantly degrade text quality across these diverse models.
\begin{table}[h]
    \centering
    \small
    \begin{tabular}{lcccc}
        \toprule
        \textbf{Model} & \textbf{TPR@1\%FPR} & \textbf{TPR@0.1\%FPR} & \textbf{WM PPL} & \textbf{Baseline PPL} \\
        \midrule
        \texttt{Olmo-3-1025-7B} & \text{1.000 $\pm$~0.000} & \text{1.000 $\pm$~0.000} & \text{19.329 $\pm$~0.377} & \text{16.801} \\
        \texttt{Qwen2.5-7B}     & \text{1.000 $\pm$~0.000} & \text{1.000 $\pm$~0.000} & \text{15.869 $\pm$~0.355} & \text{13.097} \\
        \texttt{SmolLM2-1.7B}   & \text{1.000 $\pm$~0.000} & \text{1.000 $\pm$~0.000} & \text{29.929 $\pm$~2.310} & \text{20.275} \\
        \texttt{phi-4}          & \text{1.000 $\pm$~0.000} & \text{0.999 $\pm$~0.002} & \text{7.128 $\pm$~0.158}  & \text{6.157} \\
        \bottomrule
    \end{tabular}
    \caption{\textbf{Generalizability across LLMs.} \sysname achieves perfect TPR@1\%FPR across all four models, demonstrating its robustness and generalizability. The difference between WM PPL and Baseline PPL indicates the impact of watermarking on the model's perplexity.}
    \label{tab:open-source-model-robustness}
\end{table}

\section{Detection under significant distribution shift}
\label{appendix:ood-detection}
Because the projection matrix $P$ and selector matrix $S$ are trained on English web text from OpenWebText~\citep{Gokaslan2019OpenWeb}, we further evaluate detectability and text quality on three datasets that represent a significant distribution shift from this training distribution: the Japanese subset of FineWeb-2~\citep{penedo2025fineweb2}, OpenWebMath~\citep{paster2023openwebmath} (mathematical web text), and the Python subset of StarCoderData~\citep{li2023starcoder}.
Table~\ref{tab:detect-ood} reports TPR@0.1\%FPR and perplexity under these settings. \sysname retains strong detectability on the Japanese and mathematical web-text datasets for both primary models.
Performance degrades only on the Python subset of StarCoderData, consistent with prior work showing that code is particularly challenging to watermark because of its low-entropy, highly constrained token distributions~\citep{lee2023codewatermark}.
The baselines degrade more severely under this shift. While Unremovable transfers better than GaussMark and KGW Distilled, it trails \sysnamenospace\ across all datasets, with the largest gap on code.
\begin{table}[ht]
    \centering
    \small
    \setlength{\tabcolsep}{2.5pt}
    \resizebox{\columnwidth}{!}{
    \begin{tabular}{llcccccc}
        \toprule
        \multirow{2}{*}{\textbf{Method}} & \multirow{2}{*}{\textbf{Model}} & \multicolumn{2}{c}{\textbf{OpenWebMath}} & \multicolumn{2}{c}{\textbf{FineWeb-2 (Ja)}} & \multicolumn{2}{c}{\textbf{StarCoderData}} \\
        \cmidrule(lr){3-4} \cmidrule(lr){5-6} \cmidrule(lr){7-8}
        & & \textbf{TPR} & \textbf{PPL} & \textbf{TPR} & \textbf{PPL} & \textbf{TPR} & \textbf{PPL} \\
        \midrule
        \multirow{2}{*}{\textbf{KGW}$^\dagger$}
        & \texttt{Llama-2-7B} & 0.748\pmerr{0.269} & 16.3\pmerr{1.0} & 0.678\pmerr{0.470} & 18.6\pmerr{0.7} & 0.491\pmerr{0.261} & 6.5\pmerr{0.3} \\
        & \texttt{Mistral-7B} & 0.596\pmerr{0.496} & 15.1\pmerr{0.2} & 0.980\pmerr{0.015} & 27.3\pmerr{1.0} & 0.414\pmerr{0.180} & 7.1\pmerr{0.7} \\
        \cmidrule{2-8}
        \multirow{2}{*}{\textbf{KGW + LLR}$^\dagger$}
        & \texttt{Llama-2-7B} & 0.998\pmerr{0.002} & 15.4\pmerr{0.5} & 0.998\pmerr{0.001} & 17.7\pmerr{1.3} & 0.915\pmerr{0.006} & 6.6\pmerr{0.3} \\
        & \texttt{Mistral-7B} & 0.988\pmerr{0.005} & 14.4\pmerr{0.7} & 1.000\pmerr{0.000} & 27.4\pmerr{0.7} & 0.868\pmerr{0.075} & 6.4\pmerr{0.2} \\
        \midrule
        \multirow{2}{*}{\shortstack[l]{\textbf{KGW}\\\textbf{Distilled}}}
        & \texttt{Llama-2-7B} & 0.247\pmerr{0.329} & 20.5\pmerr{1.1} & 0.247\pmerr{0.222} & 25.5\pmerr{0.0} & 0.133\pmerr{0.100} & 8.3\pmerr{0.3} \\
        & \texttt{Mistral-7B} & 0.304\pmerr{0.505} & 23.4\pmerr{1.4} & 0.554\pmerr{0.255} & 50.0\pmerr{4.4} & 0.195\pmerr{0.156} & 15.4\pmerr{1.4} \\
        \cmidrule{2-8}
        \multirow{2}{*}{\textbf{GaussMark}}
        & \texttt{Llama-2-7B} & 0.478\pmerr{0.142} & 17.3\pmerr{1.1} & 0.957\pmerr{0.031} & 25.9\pmerr{1.4} & 0.294\pmerr{0.113} & 7.8\pmerr{0.2} \\
        & \texttt{Mistral-7B} & 0.282\pmerr{0.051} & 16.4\pmerr{0.7} & 0.785\pmerr{0.193} & 36.3\pmerr{3.7} & 0.186\pmerr{0.092} & 7.3\pmerr{0.5} \\
        \cmidrule{2-8}
        \multirow{2}{*}{\textbf{Unremovable}}
        & \texttt{Llama-2-7B} & 0.912\pmerr{0.059} & 16.0\pmerr{2.3} & 0.929\pmerr{0.039} & 15.6\pmerr{2.4} & 0.720\pmerr{0.126} & 7.0\pmerr{1.5} \\
        & \texttt{Mistral-7B} & 0.855\pmerr{0.072} & 13.6\pmerr{1.8} & 0.921\pmerr{0.075} & 24.6\pmerr{1.8} & 0.599\pmerr{0.133} & 5.8\pmerr{0.3} \\
        \cmidrule{2-8}
        \multirow{2}{*}{\textbf{\sysnamenospace}}
        & \texttt{Llama-2-7B} & \textbf{0.993\pmerr{0.009}} & 14.5\pmerr{1.2} & \textbf{1.000\pmerr{0.000}} & 19.6\pmerr{3.0} & \textbf{0.923\pmerr{0.047}} & 6.9\pmerr{0.7} \\
        & \texttt{Mistral-7B} & \textbf{0.982\pmerr{0.007}} & 14.7\pmerr{0.2} & \textbf{1.000\pmerr{0.000}} & 25.3\pmerr{3.0} & \textbf{0.842\pmerr{0.088}} & 6.9\pmerr{0.4} \\
        \bottomrule
    \end{tabular}
    }
    \caption{
        \textbf{Detection under significant distribution shift.}
        We report TPR@0.1\%FPR (labeled TPR) and perplexity (PPL) on OpenWebMath~\citep{paster2023openwebmath}, the Japanese subset of FineWeb-2~\citep{penedo2025fineweb2}, and the Python subset of StarCoderData~\citep{li2023starcoder}. Bold values indicate the highest TPR among open-weight watermarking methods for each dataset and model.
    }
    \label{tab:detect-ood}
\end{table}

\section{Effect of varying \(\gamma\) and \(\delta\)}
\label{appendix:gamma-delta-effect}
\begin{figure}[h]
    \centering
    \begin{subfigure}{0.48\textwidth}
        \includegraphics[width=\linewidth]{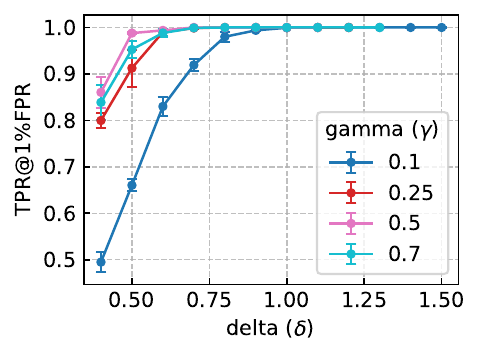}
        \caption{\textbf{Effect of \(\gamma\) and \(\delta\) on detectability.}
            TPR@1\%FPR vs.\ \(\delta\) for various \(\gamma\) values.}
        \label{fig:tpr_vs_delta}
    \end{subfigure}
    \hfill
    \begin{subfigure}{0.48\textwidth}
        \includegraphics[width=\linewidth]{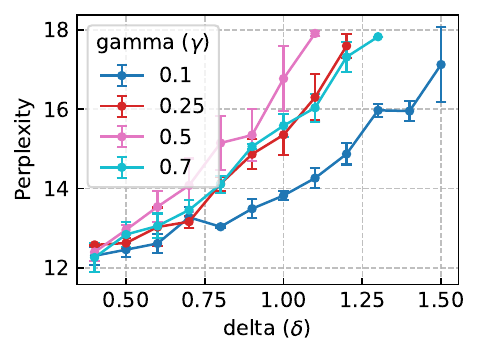}
        \caption{\textbf{Effect of \(\gamma\) and \(\delta\) on text quality.}
            Perplexity vs.\ \(\delta\) across the same \(\gamma\) settings, measured using \texttt{Llama-2-13b}.}
        \label{fig:ppl_vs_delta}
    \end{subfigure}
\end{figure}

We measure the detectability and text quality of watermarked samples generated using different values of \(\gamma\) and \(\delta\) on \texttt{Llama-2-7B}. The results, shown in Figures~\ref{fig:tpr_vs_delta} and \ref{fig:ppl_vs_delta}, indicate the trade-off between detectability and distortion as we vary these parameters.

\section{Training the projection matrix $P$}
\label{appendix:projection-training}

\begin{figure}[t]
    \centering
    \includegraphics[width=0.65\linewidth]{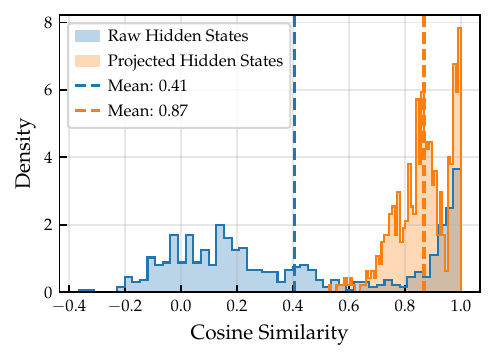}
    \caption{\textbf{Effectiveness of Projection Matrix \(P\).} The distribution of cosine similarities between original and paraphrased text pairs is shown for both raw hidden states and projected hidden states. The projection significantly increases the similarity, indicating that \(P\) successfully aligns hidden states with semantic similarity.}
    \label{fig:projection-effectiveness}
\end{figure}

The projection matrix $P$ is trained on a paired dataset of hidden states and high-quality sentence embeddings extracted from a large-scale text corpus. Specifically, hidden states are extracted from the model's final layer, while target sentence embeddings are obtained using the \texttt{Qwen3-Embedding-8B} embedding model~\citep{zhang2025qwen3}. Training utilizes $1.5$ million paired samples extracted from the OpenWebText corpus, with a maximum sequence length of $512$ tokens. 

We optimize $P$ using a contrastive loss function \ref{eq:projection-loss} designed to align the similarity distribution of projected hidden states with that of the target sentence embeddings. The optimization is performed using \texttt{AdamW}~\citep{adamw} with a learning rate of $10^{-5}$, a weight decay of $10^{-2}$, and a temperature $T_{\mathrm{sim}} = 0.1$ over $15$ epochs.

To evaluate the projection's effectiveness, we measure the cosine similarity between the projected hidden states of $500$ text samples and their corresponding paraphrases (generated via the technique described in Appendix~\ref{appendix:paraphrasing-attack}). As illustrated in Figure \ref{fig:projection-effectiveness}, the projection significantly enhances the similarity between original and paraphrased pairs compared to raw hidden states, confirming that $P$ successfully projects hidden states into a semantically aligned space where paraphrases are more closely clustered.

\paragraph{Effect of the embedding model}

The choice of sentence-embedding model used to supervise $P$ is a design decision underlying the paraphrasing robustness of \sysnamenospace. Our main experiments use \texttt{Qwen3-Embedding-8B}~\citep{zhang2025qwen3} as the target embedding space. To examine how sensitive the method is to this choice, we retrain $P$ on \texttt{Llama-2-7B} using two additional embedding models---\texttt{multilingual-e5-large}~\citep{wang2024multilinguale5} and \texttt{bge-m3}~\citep{chen2024bgem3}. We then re-evaluate paraphrasing robustness under the protocol of Section~\ref{subsec:paraphrasing}, reporting AUROC and TPR@1\%FPR on paraphrased text.

Table~\ref{tab:embedding-model-ablation} shows comparable paraphrasing robustness across all three embedding models, with each outperforming the ablation that omits semantic alignment. Under this setup, the benefit of alignment does not appear to depend on a specific embedding model.
\begin{table}[h]
    \centering
    \small
    \begin{tabular}{lcc}
        \toprule
        \textbf{Embedding Model} & \textbf{AUROC} & \textbf{TPR@1\%FPR} \\
        \midrule
        \sysname w/o sem-align & \text{0.971 $\pm$~0.005} & \text{0.762 $\pm$~0.024} \\
        \midrule
        \texttt{Qwen3-Embedding-8B}   & \textbf{0.990 $\pm$~0.003} & \textbf{0.907 $\pm$~0.003} \\
        \texttt{multilingual-e5-large} & \text{0.988 $\pm$~0.003} & \text{0.877 $\pm$~0.018} \\
        \texttt{bge-m3}               & \text{0.989 $\pm$~0.005} & \text{0.861 $\pm$~0.055} \\
        \bottomrule
    \end{tabular}
    \caption{\textbf{Effect of the embedding model used to train the projection matrix $P$.} Paraphrasing robustness on \texttt{Llama-2-7B}. The \sysname w/o sem-align row removes the projection matrix and trains the selector on raw hidden states.}
    \label{tab:embedding-model-ablation}
\end{table}

\section{Training the selector matrix}
\label{appendix:selector-training}

We train the selector matrix $S$
to classify projected hidden state vectors
into $L$ distinct classes,
where $L$ is a hyperparameter.

We apply incremental K-Means clustering on the projected hidden states and group them
into $L$ clusters. Clusters with fewer than $10$ hidden states
are discarded to ensure sufficient representation.
We set the ridge regression regularization parameter \(\lambda\) to \(10^{-3}\)
throughout our experiments.

\section{Paraphrasing attack setup}
\label{appendix:paraphrasing-attack}
We generate paraphrased variants of the watermarked completions by prompting \texttt{Qwen2.5-14B-Instruct}~\citep{qwen2.5} using the template provided below. For generation, we employ nucleus sampling with $p = 0.9$, a temperature of $0.7$, and beam search with a beam width of $3$.

\begin{promptbox}{Prompt Template: Paraphrase Task}
Instruction: The input contains a Prompt and its Completion. Paraphrase ONLY the Completion. Preserve the exact meaning, facts, scope, and intent. Do not add, remove, infer, or reinterpret any information. Change wording and sentence structure only.

\bigskip

Prompt:\\
\{\{prompt\}\}

\bigskip

Completion:\\
\{\{completion\}\}

\bigskip

Paraphrased Completion:
\end{promptbox}

\begin{figure}[h]
    \centering
    \includegraphics[width=0.65\linewidth]{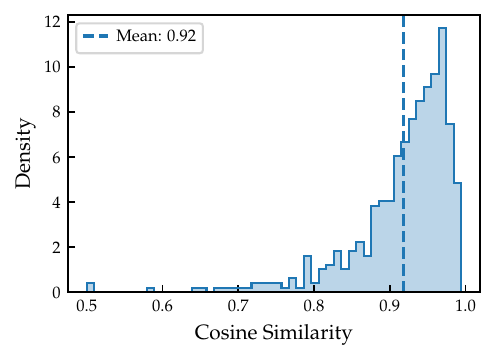}
    \caption{\textbf{Quality of Paraphrased Completions.} The distribution of cosine similarities between original and paraphrased completions shows a high mean similarity of 0.92, indicating that the paraphrasing attack successfully preserves the semantic content of the original completions.}
    \label{fig:paraphrase-quality}
\end{figure}

To assess paraphrase quality, we compute the cosine similarity between original and paraphrased outputs for 500 samples using the \texttt{Qwen3-Embedding-8B} embedding model. As illustrated in Figure \ref{fig:paraphrase-quality}, the similarity scores are clustered around a mean of 0.92, confirming that the paraphrases maintain high semantic fidelity to the original completions.

\section{Hyperparameter details for watermarking methods}
\label{appendix:experimental-setup}

This section details the experimental setup and hyperparameters used across all experiments. For each watermarking method, we perform a parameter sweep to generate the plot in Figure~\ref{fig:ppl_tpr_plot}. For other evaluations, such as robustness to paraphrasing attacks (\S~\ref{subsec:paraphrasing}), downstream task accuracy (\S~\ref{subsec:downstream}), and resistance to post-hoc model modifications (\S~\ref{subsec:durability}), we select a single representative configuration from each method.

\paragraph{\sysname}
We encode $L = 254$ green lists, each containing a fraction $\gamma = 0.25$ of green tokens for both \texttt{Llama-2-7B} and \texttt{Mistral-2-7B}. The strength parameter $\delta$ is selected from the interval $[0.3, 1.2]$ for the Pareto evaluation. For all other experiments, we select $\delta = 1.0$ as the representative configuration.

\paragraph{GaussMark baseline}
For GaussMark~\citep{block2025gaussmark}, we perturbed the MLP up-projection weights in a single decoder block using Gaussian noise. For \texttt{Llama-2-7b}, the 27th decoder block was used with $\sigma \in \{0.025, 0.03, 0.035, 0.04, 0.045\}$ for the Pareto evaluation. The configuration $\sigma = 0.04$ is chosen for all other experiments. For \texttt{Mistral-2-7b}, the 20th decoder block and $\sigma = 0.005$ was selected for evaluations.

\paragraph{Unremovable baseline}
For Unremovable~\citep{christ2024provably}, the strength parameter $\varepsilon$ is selected from $\{0.3, 0.4, \dots, 0.9\}$ for the Pareto evaluation on \texttt{Llama-2-7B}. For all other experiments on both \texttt{Llama-2-7B} and \texttt{Mistral-7B}, we select $\varepsilon = 0.8$ as the representative configuration.

\paragraph{KGW distilled baseline}
We trained several logit-distilled KGW variants~\citep{gu2023learnability} of \texttt{Llama-2-7B} using a green list fraction $\gamma = 0.25$ and strength values $\delta \in \{1.0, 1.25, 1.5, 1.75, 2.0\}$ to construct the Pareto frontier. The distilled models with $\delta = 2.0$ was selected for all other evaluations.

\paragraph{KGW decoding-time watermark}
We used the decoding-time KGW watermark~\citep{kirchenbauer2023watermark}, which biases the model's logits during generation without requiring any parameter modification. We set $k=1$, which is the token context length used by the PRF to generate green lists. We fixed the green list fraction $\gamma = 0.25$ and swept the bias strength $\delta \in \{0.7, 0.8, \dots, 2.0\}$ to populate the Pareto frontier. For other evaluations, we selected $\delta = 1.5$ as the representative configuration.

\paragraph{Semantic Invariant Robust (SIR) baseline}
We evaluate SIR~\citep{liu2024a} as a decoding-time reference for paraphrasing robustness, with bias strength $\delta = 1.0$ and chunk length $10$ on both models.

\paragraph{KGW + LLR detection variant}
We additionally evaluated KGW with an LLR-based detection strategy (Section~\ref{subsec:watermark-detection}) to simulate a white-box detection setting. The hyperparameters were fixed at $\gamma = 0.25$ and $\delta \in \{0.5, 0.6, \dots, 1.7\}$ for the Pareto evaluation. For other evaluations, we selected $\delta = 1.4$.

\section{Practical challenges with RL-based watermarking}
\label{appendix:rl-watermarking-challenges}
We attempted to implement the RL-based watermarking method from \citet{xu2025learning}. However, the trained RL model was unable to generate detectable watermarks when the text was generated using multinomial sampling, and it could only embed a strong watermark when sampling with greedy decoding. This is not practical for real-world applications, as greedy decoding often produces repetitive and low-quality text.

We quantify repetition using seq-rep-3, the proportion of duplicate 3-grams in a sequence ~\citep{welleck2020neural}:
\[
    1 - \frac{\# \text{ of unique 3-grams}}{\# \text{ of 3-grams}}
\]
We report mean \texttt{seq-rep-3} across watermarked samples for both RL watermarking and our method under greedy and multinomial (temperature = 1.0) decoding, along with TPR@1\%FPR. Results in Table~\ref{tab:repetition} show that while greedy decoding makes RL watermarks detectable, it also causes severe repetition, limiting practicality. In contrast, our method preserves low repetition while maintaining strong detectability across both decoding strategies.

\begin{table}[h]
    \centering
    \begin{tabular}{lcccc}
        \toprule
        \multirow{2}{*}{\textbf{Method}} & \multicolumn{2}{c}{\textbf{Multinomial}} & \multicolumn{2}{c}{\textbf{Greedy}}                                            \\
        \cmidrule(lr){2-3} \cmidrule(lr){4-5}
                                         & \textbf{TPR@1\%FPR}                      & \textbf{seq-rep-3}                  & \textbf{TPR@1\%FPR} & \textbf{seq-rep-3} \\
        \midrule
        RL Watermarking                  & 0.25                                     & 0.03                                & 0.99                & 0.58               \\
        \sysnamenospace                  & 1.0                                      & 0.03                                & 1.0                 & 0.03               \\
        \bottomrule
    \end{tabular}
    \caption{\textbf{Detectability (TPR@1\%FPR) and text repetition (mean seq-rep-3) for RL watermarking and \sysname\ under multinomial and greedy decoding.} While RL watermarking achieves high detectability with greedy decoding, it causes severe repetition, whereas \sysname\ maintains both strong detectability and low repetition across settings.}

    \label{tab:repetition}
\end{table}

\section{Fine-tuning setup details}
\label{appendix:finetune-details}

We fine-tune all models on OpenWebText. We follow a setup similar to ~\cite{gloaguen2025towards}: we use a batch size of 48 with 512 tokens per input and a learning rate of $2e^{-5}$. We use the Adafactor optimizer with cosine learning rate decay and linear warmup for the first 500 steps. For LoRA~\citep{hu2022lora}, we set the rank to 16, the scaling factor (alpha) to 32, and the dropout rate to $0.1$. We fine-tune all the internal linear layers of the transformer blocks along with the unembedding layer.

For instruction fine-tuning, we instead train for one epoch on Alpaca~\citep{taori2023alpaca} (\texttt{tatsu-lab/alpaca}) with effective batch size 128 and a maximum sequence length of 512. We use the same optimizer, learning rate, and LoRA hyperparameters as above, but apply LoRA only to attention and MLP projections, corresponding to a realistic post-release instruction-tuning setup.

\section{Analysis of watermarking behavior}
\label{appendix:understanding-watermark-behavior}

In this section, we study how the hyperparameter \(L\), which represents the number of green lists, influences our watermarking method. We focus on two aspects: (1) the alignment between watermark logits and the green list structure, and (2) the detection performance.

\begin{figure*}[t]
  \centering
  \begin{minipage}[t]{0.48\textwidth}
    \centering
    \includegraphics[width=\linewidth]{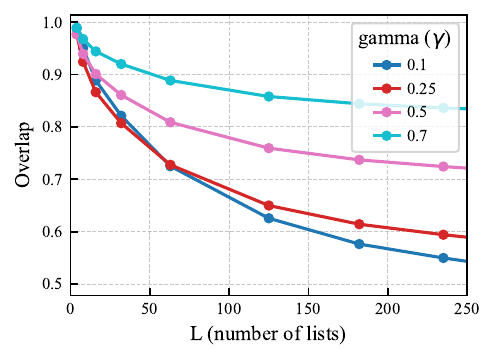}
    \caption{\textbf{Green List Token Overlap.}
      As \(L\) increases, the overlap between the top-biased tokens and the intended green list decreases, indicating weakening alignment with the intended green list structure.
    }
    \label{fig:overlap-vs-L}
  \end{minipage}\hfill
  \begin{minipage}[t]{0.48\textwidth}
    \centering
    \includegraphics[width=\linewidth]{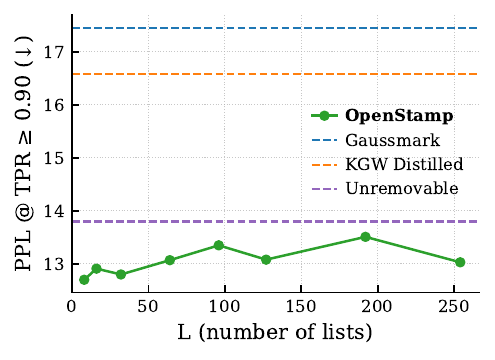}
    \caption{\textbf{Detection Performance vs. \(L\).}
      Minimum PPL to achieve TPR@1\%FPR $\geq 0.90$ is reported for different \(L\) values. PPL remains stable with growing \(L\) and consistently below the GaussMark, KGW Distilled, and Unremovable baselines.
    }
    \label{fig:detection-vs-L}

  \end{minipage}
\end{figure*}

\paragraph{Measuring alignment.}
We first extract 5{,}000 hidden states from OpenWebText samples. For each hidden state \(h\), we compute watermark logits \(\Delta W h = G S h\). We measure \emph{overlap}, defined as the fraction of
\(\lvert\mathcal{G}\rvert = \gamma \cdot \lvert\mathcal{V}\rvert\) tokens
with the largest logits in \(\Delta W h\) that also belong to the intended green list
\(\mathcal{G}_{\ell}\), where \(\ell = \arg\max_i (Sh)_i\). We report overlap for different values of \(\gamma\) and \(L\) in Figure~\ref{fig:overlap-vs-L}. As \(L\) increases, the overlap decreases, indicating a weakening alignment with the green list structure.

\paragraph{Measuring detection performance.}
We find the minimum PPL required to achieve TPR@1\%FPR $\geq 0.90$ on watermarked samples. A lower PPL threshold indicates a more effective watermark, since it enables reliable detection with less degradation in text quality. We report the PPL threshold for different \(L\) values. For comparison, we also report corresponding PPL thresholds for GaussMark, KGW Distilled, and Unremovable. Figure~\ref{fig:detection-vs-L} shows that PPL remains stable across different $L$ values and consistently below the baseline thresholds.

\section{Effect of \(L\) on logit variability}
\label{appendix:variability-watermark-logits}
\begin{figure}[h]
    \centering
    \includegraphics[width=0.6\linewidth]{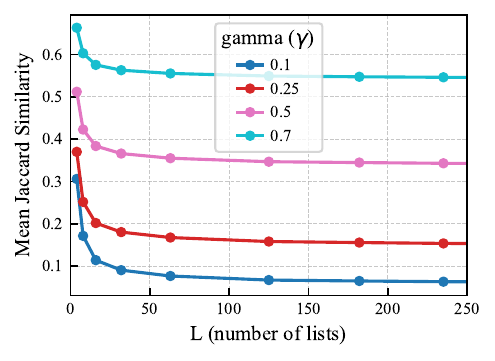}
    \caption{\textbf{Variability in Watermark Logits.}
        As \(L\) increases, the mean Jaccard similarity between different top-biased token sets decreases, indicating greater variability in watermark logits across different contexts.
    }
    \label{fig:variability-vs-L}
\end{figure}

We study how increasing \(L\) affects the variability of watermark logits across different contexts. With more green lists available, hidden states can be assigned to a greater number of distinct green lists. This increases the diversity of tokens favored by the watermark, making it less predictable and more robust to reverse-engineering attacks. We measure variability by computing the mean Jaccard similarity between the sets of token indices corresponding to the top $\gamma|V|$ components of watermark logits across hidden states. We evaluate this metric for different values of $L$ and $\gamma$. Lower similarity implies greater variability in the watermark logits. As shown in Figure~\ref{fig:variability-vs-L}, mean similarity decreases with increasing \(L\), though the decline plateaus for large \(L\), reflecting diminishing marginal gains in variability.

\section{LLM usage}
\label{appendix:llm-usage}
ChatGPT~\citep{openai_chatgpt} was used to assist with typesetting (e.g., equations, tables) and for minor language editing (grammar and conciseness). GitHub Copilot~\citep{github_copilot} was used to assist in implementing the code for aggregating results and generating plots. All outputs were reviewed and validated by the authors.

\section{Prompt independence of watermark detection}
\label{appendix:prompt-independence}
\begin{figure}
    \centering
    \begin{subfigure}{\textwidth}
        \centering
        \includegraphics[width=\linewidth]{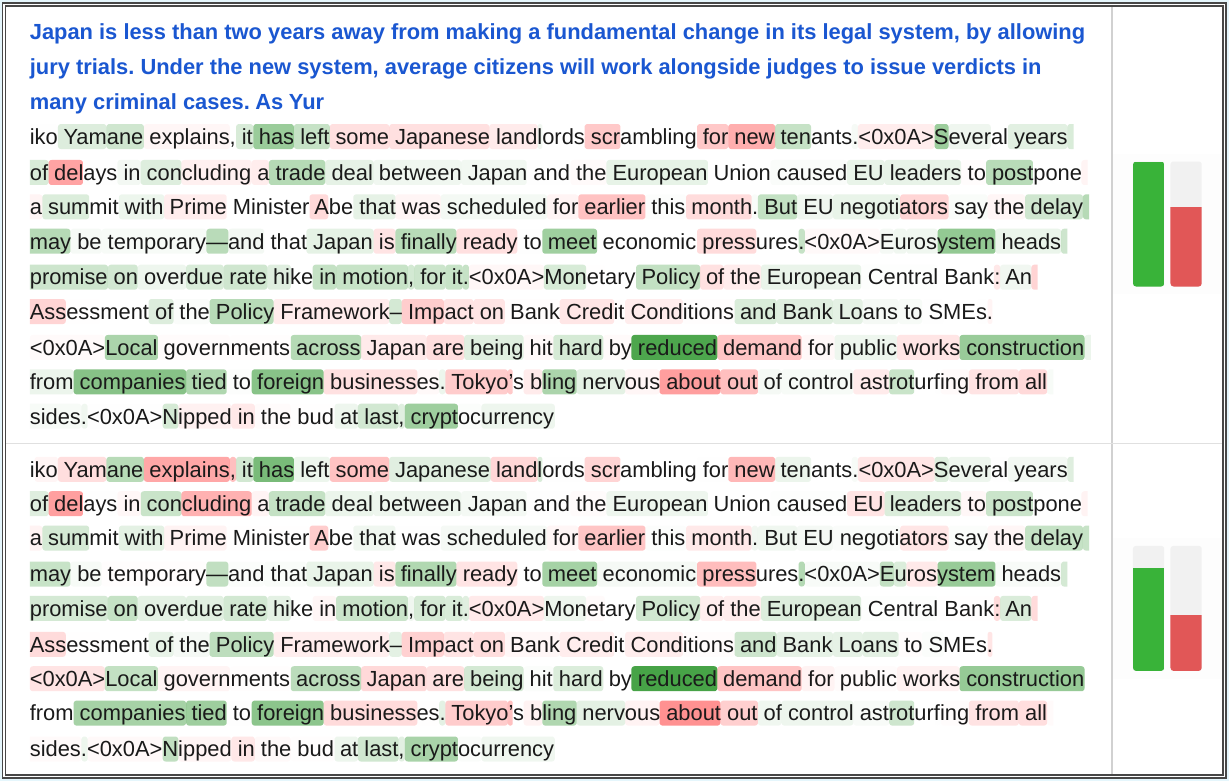}
        \caption{Watermarked Text}
        \label{fig:prompt-independence-watermarked}
    \end{subfigure}

    \vspace{1em}

    \begin{subfigure}{\textwidth}
        \centering
        \includegraphics[width=\linewidth]{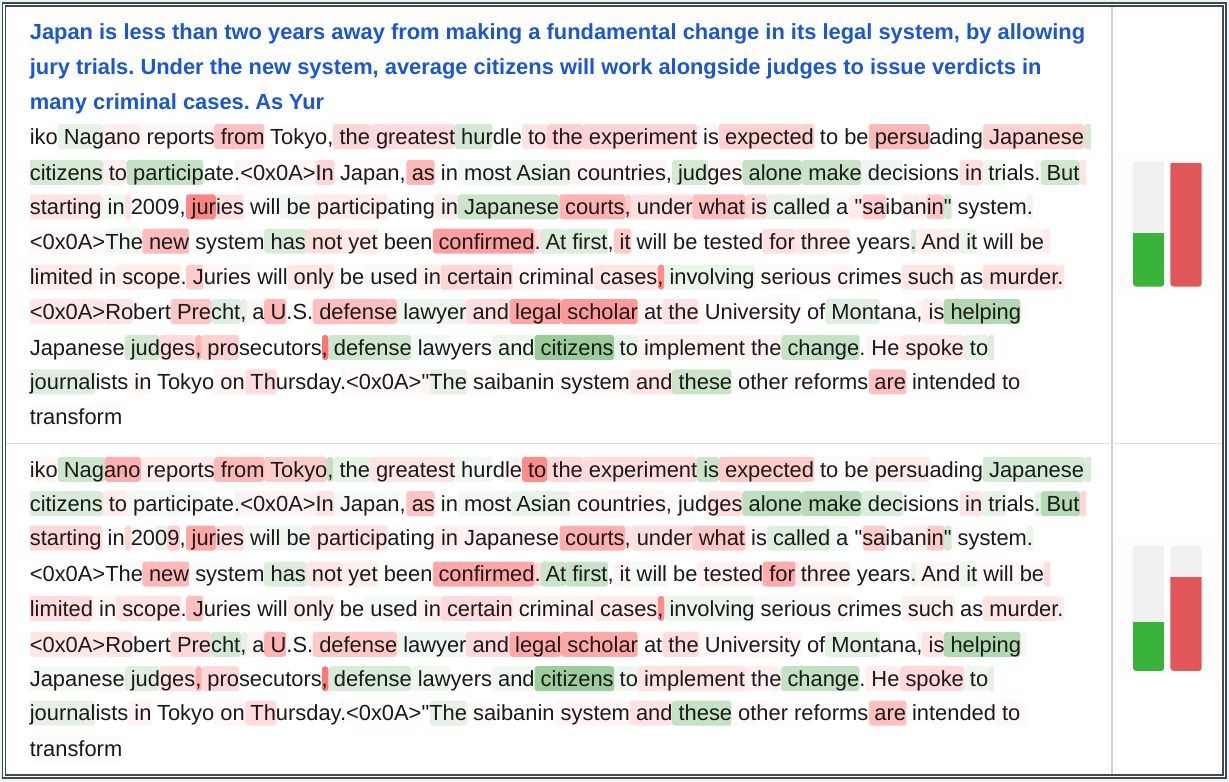}
        \caption{Unwatermarked Text}
        \label{fig:prompt-independence-unwatermarked}
    \end{subfigure}

    \caption{\textbf{Prompt Independence of Watermark Detection.} Per-token LLR scores for (a) watermarked and (b) unwatermarked text, computed with and without the prompt. The red highlight indicates tokens with a negative score whereas the green highlight indicates tokens with a positive score.
        The text in blue is the prompt. Scores are similar in both settings, indicating that detection is robust to the absence of the original prompt.}
    \label{fig:prompt-independence}
\end{figure}

To explain why our detection method works without the original prompt, we compare per-token LLR scores on watermarked and unwatermarked text, with and without the prompt. As shown in Figures~\ref{fig:prompt-independence-watermarked} and \ref{fig:prompt-independence-unwatermarked}, the scores are largely consistent across settings, diverging only slightly at the start of generation. This suggests that hidden states—and thus green-list selections—depend mainly on the immediate local context rather than the initial prompt, allowing effective watermark detection even when the prompt is unavailable.

\section{Additional attack setups}
\label{appendix:additional-finetuning-attacks}

Here we evaluate a selector-inversion attack aimed at reconstructing the selector matrix $S$.

\subsection{Inversion attack via reconstructing the selector matrix}

We evaluate a plausible inversion attack based on the intuition that an adversary might try to reconstruct the selector matrix S and then reverse-engineer each cluster's green list. In practice, this attack is particularly challenging because key details—such as the number of clusters L, the k-means initialization seed, and the dataset used to extract hidden states—are kept private by the model provider.

Even if the attacker guesses L, reproducing the original hidden-state-to-cluster assignments is highly sensitive to both the k-means seed and the underlying dataset. To illustrate this sensitivity, we construct selector matrices with the same \(L\) using different datasets (OpenWebText vs.\ FineWeb) and different initialization seeds. We then generate a test set of hidden states from RealNewsLike samples and compute cluster assignments under each variant of the selector matrix. Agreement between assignments is measured using Adjusted Rand Index (ARI) and Normalized Mutual Information (NMI). As shown in Table~\ref{tab:selector-reconstruction}, cluster assignments vary across datasets and seeds. This variability suggests that reconstructing the original selector matrix is challenging, providing inherent robustness against this class of inversion attacks.

\begin{table}[h]
    \centering
    \vspace{0.5em}
    \begin{tabular}{l c c}
        \toprule
        Condition                 & ARI  & NMI  \\
        \midrule
        Different Dataset         & 0.44 & 0.65 \\
        Different Seed            & 0.46 & 0.67 \\
        Different Dataset \& Seed & 0.37 & 0.63 \\
        \bottomrule
    \end{tabular}
    \caption{\textbf{Agreement between cluster assignments from different selector matrices.} The ARI and NMI scores indicate moderate agreement, showing that reconstructing the original selector matrix is challenging even with knowledge of \(L\).}
    \label{tab:selector-reconstruction}
\end{table}

\section{LLR vs. discretized-signal detectors}
\label{appendix:freq-ablation}
We assess how much the LLR detector benefits from retaining the full continuous watermark signal versus \emph{discretizing} it.
For this, we compare the LLR detector against two
ablations that discretize the watermark signal: (i) a discrete LLR variant that
replaces the soft selector with an argmax before computing likelihoods, and (ii) a
binomial count detector that counts the number of selected green list tokens,
following the style of \citet{kirchenbauer2023watermark}. In the discrete LLR variant, the selector $s = S h_t$ is collapsed to a single index
$\hat{\ell}_t = \arg\max_i s_i$, and the corresponding green list
vector $G_{\hat{\ell}_t} \in \mathbb{R}^{|V|}$ is taken as the discretized signal.
The watermarked distribution is then
\[
  p_{\mathrm{wm}}(x_t \mid G_{\hat{\ell}_t})
  = \frac{\exp\!\bigl(v_t + G_{\hat{\ell}_t}\bigr)_{x_t}}
  {\sum_{w\in V} \exp\!\bigl(v_t + G_{\hat{\ell}_t}\bigr)_w},
\]
which the mirrors the numerator in Equation~\ref{eq:llr} but uses only a single green-list vector, removing all
mixed-list contributions present in the full model. The binomial count detector instead treats green list membership as a Bernoulli
indicator, forming
\[
  Z = \sum_{t=1}^T \mathbbm{1}\{x_t \in G_{\hat{\ell}_t}\},
  \qquad
  z = \frac{Z - \gamma T}{\sqrt{T\gamma(1-\gamma)}}.
\]

Table \ref{tab:detector-ablation} shows TPR at various FPR thresholds for all three detectors. The two discretized-signal variants recover some watermark signal but they still perform noticeably worse than the LLR detector, indicating that collapsing the mixed green list structure into a single discrete choice loses information the continuous LLR leverages.
\begin{table}[h]
\centering
\begin{tabular}{lccc}
\toprule
\textbf{Method} & \textbf{TPR@1\%FPR} & \textbf{TPR@0.1\%FPR} & \textbf{TPR@0.001\%FPR} \\
\midrule
Discrete LLR    & 0.96 & 0.82 & 0.79 \\
Binomial Count  & 0.86 & 0.51 & 0.50 \\
\textbf{LLR}             & \textbf{1.00} & \textbf{1.00} & \textbf{1.00} \\
\bottomrule
\end{tabular}
\caption{\textbf{Comparing LLR to alternative detectors}. The discretized-signal variants perform worse than the full LLR detector, indicating that collapsing the mixed green list structure into a single discrete choice loses useful information.}
\label{tab:detector-ablation}
\end{table}

\section{Effect of the LLR detector on baselines}
\label{appendix:llr-baselines}
To isolate the effect of the LLR detector, we rescore GaussMark, KGW Distilled, and Unremovable with the length-normalized LLR in Equation~\ref{eq:llr} instead of each method's native detector.
The generated texts remain unchanged, so any change in TPR@0.1\%FPR is attributable solely to the choice of detector.

Table~\ref{tab:llr-openweight-baselines} shows that the LLR detector improves detectability for all three methods on both \texttt{Llama-2-7B} and \texttt{Mistral-7B}, with the largest gain observed for GaussMark.
These results suggest that, when white-box access to the model is available, the LLR detector can provide more effective detection.
\begin{table}[h]
\centering
\small
\begin{tabular}{llcc}
\toprule
\textbf{Method} & \textbf{Model} & \textbf{Native} & \textbf{LLR} \\
\midrule
\multirow{2}{*}{\textbf{GaussMark}}
& \texttt{Llama-2-7B} & 0.737$\pm$0.089 & 0.922$\pm$0.017 \\
& \texttt{Mistral-7B} & 0.349$\pm$0.121 & 0.566$\pm$0.155 \\
\cmidrule{2-4}
\multirow{2}{*}{\textbf{KGW Distilled}}
& \texttt{Llama-2-7B} & 0.965$\pm$0.023 & 1.000$\pm$0.000 \\
& \texttt{Mistral-7B} & 0.996$\pm$0.007 & 1.000$\pm$0.000 \\
\cmidrule{2-4}
\multirow{2}{*}{\textbf{Unremovable}}
& \texttt{Llama-2-7B} & 0.966$\pm$0.016 & 1.000$\pm$0.000 \\
& \texttt{Mistral-7B} & 0.979$\pm$0.011 & 1.000$\pm$0.000 \\
\bottomrule
\end{tabular}
\caption{\textbf{Effect of the LLR detector on baselines} Replacing each method's native detector with the LLR improves TPR@0.1\%FPR on both models, with the largest gain for GaussMark.}
\label{tab:llr-openweight-baselines}
\end{table}

\end{document}